\documentclass[conference]{IEEEtran}
\IEEEoverridecommandlockouts

\usepackage{cite}
\usepackage{amsmath, amssymb, amsfonts}
\usepackage{algorithmic}
\usepackage{graphicx}
\usepackage{textcomp}
\usepackage{xcolor}
\usepackage{url}
\usepackage{booktabs}
\usepackage{subfigure}
\usepackage{hyperref}

\def\BibTeX{{\rm B\kern-.05em{\sc i\kern-.025em b}\kern-.08em T\kern-.1667em\lower.7ex\hbox{E}\kern-.125emX}}
\IEEEoverridecommandlockouts
\IEEEpubid{\makebox[\columnwidth]{979-8-3315-8654-6/25/\$31.00~\copyright2025 IEEE \hfill}
\hspace{\columnsep}\makebox[\columnwidth]{ }}

\begin{document}
    \title{YOLO and SGBM Integration for Autonomous Tree Branch Detection and
    Depth Estimation in Radiata Pine Pruning Applications}

    \author{\IEEEauthorblockN{Yida Lin, Bing Xue, Mengjie Zhang} \IEEEauthorblockA{\small \textit{Centre for Data Science and Artificial Intelligence} \\ \textit{Victoria University of Wellington, Wellington, New Zealand}\\ linyida\texttt{@}myvuw.ac.nz, bing.xue\texttt{@}vuw.ac.nz, mengjie.zhang\texttt{@}vuw.ac.nz}
    \and \IEEEauthorblockN{Sam Schofield, Richard Green} \IEEEauthorblockA{\small \textit{Department of Computer Science and Software Engineering} \\ \textit{University of Canterbury, Canterbury, New Zealand}\\ sam.schofield\texttt{@}canterbury.ac.nz, richard.green\texttt{@}canterbury.ac.nz}
    }
     \maketitle
    \IEEEpubidadjcol

    \begin{abstract}
        Manual pruning of radiata pine trees poses significant safety risks due to
        extreme working heights and challenging terrain. This paper presents a
        computer vision framework that integrates YOLO object detection with
        Semi-Global Block Matching (SGBM) stereo vision for autonomous drone-based
        pruning operations. Our system achieves precise branch detection and depth
        estimation using only stereo camera input, eliminating the need for
        expensive LiDAR sensors. Experimental evaluation demonstrates YOLO's superior
        performance over Mask R-CNN, achieving 82.0\% mAP$_{mask50\text{--}95}$
        for branch segmentation. The integrated system accurately localizes branches
        within a 2-meter operational range with processing times under one second
        per frame. These results establish the feasibility of cost-effective
        autonomous pruning systems that enhance worker safety and operational
        efficiency in commercial forestry.
    \end{abstract}


    \section{Introduction}

    Radiata pine (\textit{Pinus radiata}) dominates New Zealand's forestry
    industry, comprising 90\% of the country's 1.7 million hectares of
    commercial plantations and generating approximately NZD 6.8 billion in
    annual export revenue \cite{van2013national,mason2023impacts}. Regular pruning
    is essential for promoting straight trunk growth and producing high-quality,
    knot-free timber for premium markets.

    Current manual pruning practices expose workers to severe occupational
    hazards. Industry statistics reveal fatality rates of 110 per 100,000 workers—approximately
    30 times higher than the national average across all industries. Non-fatal
    injury rates reach 239 per 10,000 workers, substantially exceeding the cross-industry
    average \cite{molina2017aerial}. These safety concerns, combined with the physically
    demanding nature of the work, create significant recruitment and retention
    challenges in the forestry sector.

    While autonomous drone systems offer a promising alternative to manual
    pruning, existing implementations face several critical limitations: (1)
    continuous human supervision requirements, (2) inability to handle branches smaller
    than 20 mm diameter, and (3) dependence on expensive LiDAR sensors costing
    tens of thousands of dollars. These constraints significantly limit the economic
    viability of drone-based pruning for commercial operations.

    This research addresses these limitations by developing a cost-effective computer
    vision system that achieves three key objectives:
    \begin{itemize}
        \item Detects and segments branches as thin as 10 mm diameter using advanced
            YOLO architectures

        \item Estimates three-dimensional branch positions using stereo vision with
            sub-centimeter accuracy

        \item Operates with processing times under one second per frame,
            suitable for autonomous navigation
    \end{itemize}

    Our approach strategically combines state-of-the-art object detection with classical
    stereo vision techniques, eliminating expensive sensor requirements while
    maintaining high accuracy. This integration makes drone-based pruning economically
    viable for widespread forestry adoption.

    \section{Related Work}

    Our framework builds upon three fundamental research areas: modern object detection
    and segmentation techniques, stereo vision depth estimation methods, and computer
    vision applications in forestry automation.

    \subsection{Object Detection and Segmentation}

    Modern object detection has evolved significantly since the introduction of
    R-CNN \cite{girshick2014rich}. The field progressed through Fast R-CNN with ROI
    Pooling \cite{girshick2015fast}, Faster R-CNN with Region Proposal Networks
    \cite{ren2016faster}, and Mask R-CNN extending to pixel-level segmentation \cite{he2017mask}.

    For real-time applications, the YOLO family has gained prominence due to its
    optimal speed-accuracy balance \cite{reis2023real,wang2024yolov9}. Recent YOLO
    versions incorporate advanced segmentation capabilities while maintaining real-time
    performance, making them particularly suitable for autonomous applications
    where processing latency directly impacts operational safety and system
    responsiveness.

    \subsection{Stereo Vision and Depth Estimation}

    Depth estimation methods broadly categorize into active and passive approaches.
    Active systems employ LiDAR or structured light sensors
    \cite{wu2018squeezeseg}, while passive techniques rely on stereo matching or
    monocular depth inference \cite{hartley2003multiple}.

    For stereo camera systems, depth is computed through triangulation based on
    disparity measurements. The fundamental relationship for a calibrated stereo
    system with baseline distance $b$ is:

    \begin{equation}
        z = \frac{b \cdot f_{x}}{d}= \frac{b \cdot f_{x}}{u_{l}- u_{r}}\label{eq:stereo_depth}
    \end{equation}

    where $f_{x}$ is the horizontal focal length, $d$ is the disparity between corresponding
    points in left $(u_{l}, v_{l})$ and right $(u_{r}, v_{r})$ images, and $z$
    is the computed depth.

    Semi-Global Block Matching (SGBM) \cite{hirschmuller2007stereo} achieves an
    optimal balance between computational efficiency and accuracy by employing
    semi-global cost aggregation across multiple directional paths. This approach
    significantly improves robustness in texture-poor regions that are common in
    forest environments, making it particularly suitable for natural scene depth
    estimation.

    \subsection{Computer Vision in Forestry}

    Recent advances have enabled significant progress in forestry automation,
    including species classification \cite{fricker2019convolutional}, trunk detection
    \cite{liu2018computer}, and inventory management \cite{tang2015automatic}. However,
    autonomous branch detection presents unique technical challenges: highly variable
    branch morphologies, complex lighting conditions under forest canopies, and stringent
    precision requirements for safe autonomous operation.

    Most existing drone-based forestry systems rely on expensive LiDAR sensors \cite{bazargani2024automation},
    creating significant cost barriers for widespread commercial adoption. Our research
    directly addresses this limitation by developing a stereo vision system that
    maintains necessary spatial accuracy while utilizing only affordable camera hardware,
    thereby reducing system costs by an order of magnitude.

    \section{Methodology}

    Our integrated approach combines advanced branch detection with precise depth
    estimation for autonomous three-dimensional branch localization. The system
    architecture consists of four sequential components: stereo data acquisition
    and preprocessing, branch detection and segmentation, depth map generation using
    stereo matching, and spatial integration for accurate 3D localization.

    \begin{figure}[htbp]
        \centering
        \caption{System architecture integrating YOLO-based branch segmentation
        with SGBM stereo matching for accurate 3D branch localization.}
        \includegraphics[width=0.8\columnwidth]{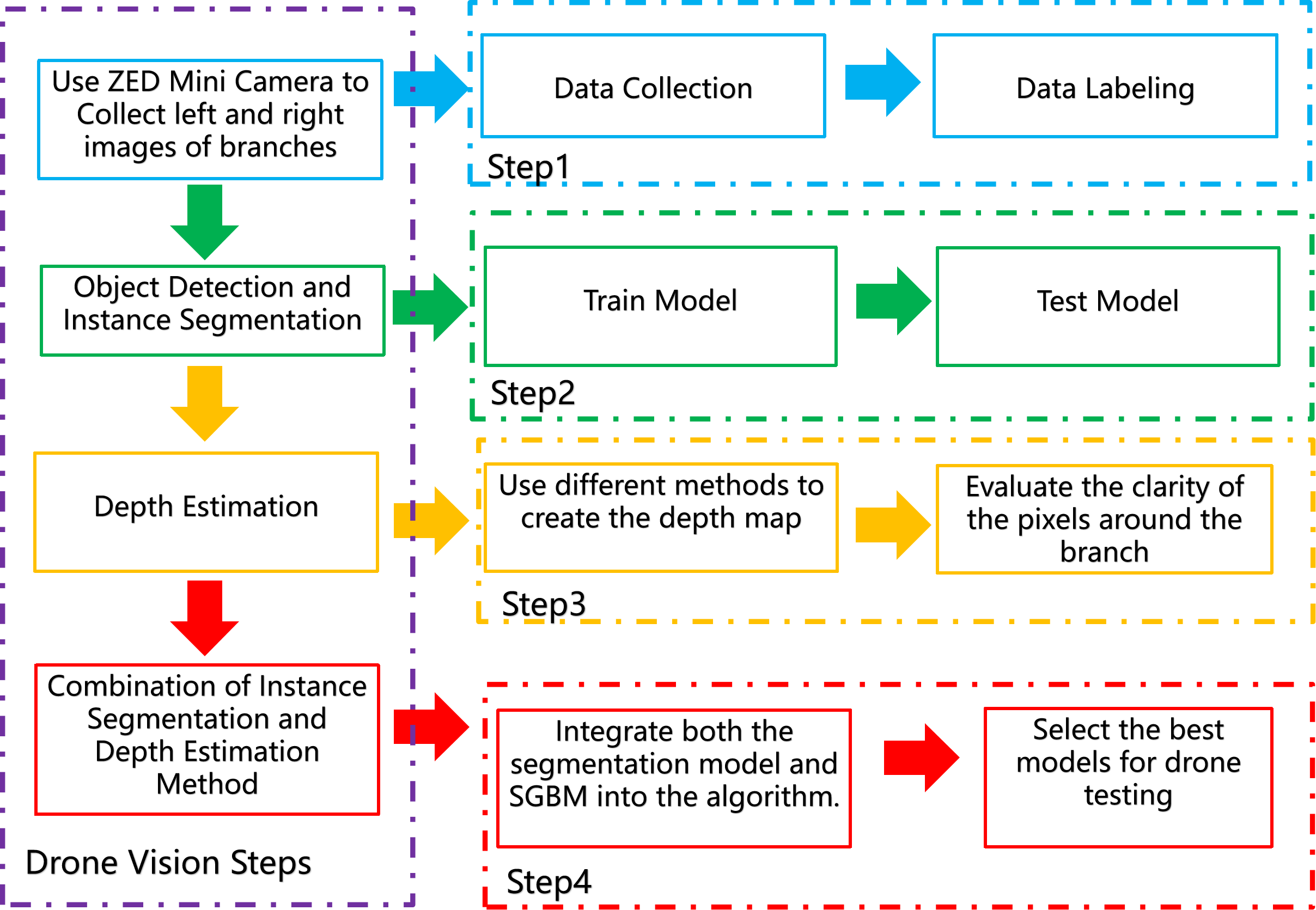}
        \label{flow_chart}
    \end{figure}

    \subsection{Data Acquisition and Preprocessing}

    Data collection utilized a ZED Mini stereo camera system capturing
    synchronized image pairs at 1920$\times$1080 resolution. We systematically collected
    branch imagery under diverse lighting scenarios and viewing angles in controlled
    laboratory environments to ensure dataset comprehensiveness. The final
    dataset comprises 61 stereo pairs for training and 10 pairs for testing, providing
    sufficient data for model development and validation.

    Manual annotation involved precise branch contour labeling using polygon boundaries
    to ensure pixel-level accuracy. To enhance model generalization and
    robustness, we implemented comprehensive data augmentation strategies including
    random rotations ($\pm$15°), horizontal flipping, and photometric
    adjustments (brightness $\pm$20\%, contrast $\pm$15\%) to simulate varying environmental
    conditions.

    \subsection{Branch Detection and Segmentation}

    We conducted a comprehensive evaluation of two distinct detection paradigms
    to determine optimal branch segmentation performance:

    \textbf{Mask R-CNN Implementation:} Multiple backbone architectures were systematically
    tested, including ResNet-50 and ResNet-101 with various configurations (C4,
    DC5, FPN), and ResNeXt-101-32x8d featuring grouped convolutions for enhanced
    feature representation.

    \textbf{YOLO Segmentation Models:} Recent YOLO versions with integrated
    segmentation capabilities were evaluated, encompassing YOLOv8 variants (nano
    through extra-large) and YOLOv9 architectures (compact and efficient) to assess
    performance across different model complexities.

    All models employed identical training protocols to ensure fair comparison:
    100 epochs with Adam optimization, initial learning rate of 0.001 with cosine
    annealing scheduling, batch size of 16, and standardized augmentation strategies.

    \subsection{Stereo Depth Estimation with SGBM}

    Depth map generation employs SGBM with carefully optimized parameters to
    maximize accuracy and robustness:
    \begin{itemize}
        \item Minimum disparity: 0

        \item Disparity range: 128 pixels

        \item Block size: 5$\times$5 pixels for local matching

        \item Smoothness penalties: $P_{1}$ = 600, $P_{2}$ = 2400 for optimal cost
            aggregation

        \item Uniqueness ratio: 10\% to filter ambiguous matches

        \item Speckle filtering: window = 100, range = 32 for noise reduction
    \end{itemize}

    Post-processing incorporates Weighted Least Squares (WLS) filtering to
    significantly refine disparity maps by preserving critical edge structures
    at branch boundaries while effectively smoothing homogeneous regions to
    reduce noise artifacts.

    \subsection{Spatial Integration and 3D Localization}

    Three-dimensional branch localization combines segmentation masks with depth
    information through a systematic pipeline:

    \begin{enumerate}
        \item Apply trained segmentation model to identify and delineate branch regions

        \item Generate high-quality disparity map using SGBM with WLS post-processing

        \item Convert disparity values to metric depth using Equation~\ref{eq:stereo_depth}

        \item Spatially register segmented pixels with corresponding depth locations

        \item Extract depth values for all pixels within detected branch segments

        \item Compute robust statistical depth measures (mean, median, std) for each
            detected branch to ensure reliable 3D localization
    \end{enumerate}

    \begin{figure}[htbp]
        \centering
        \caption{Spatial integration process: (a) Branch regions detected by YOLO
        segmentation, (b) Corresponding SGBM depth map enabling precise 3D
        branch localization.}
        \subfigure[YOLO Detection]{\includegraphics[width=0.23\textwidth]{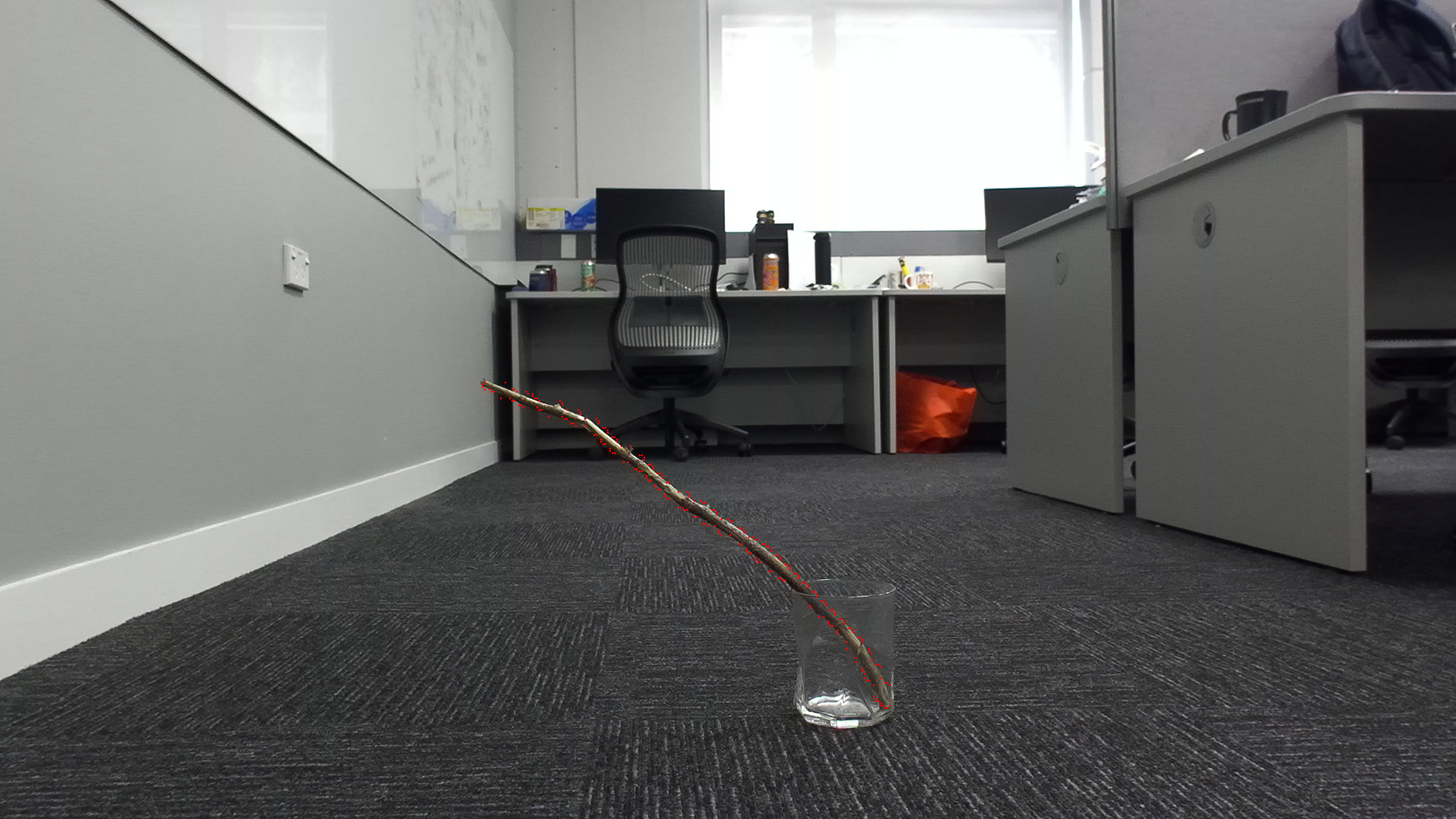}}
        \subfigure[SGBM Depth Map]{\includegraphics[width=0.23\textwidth]{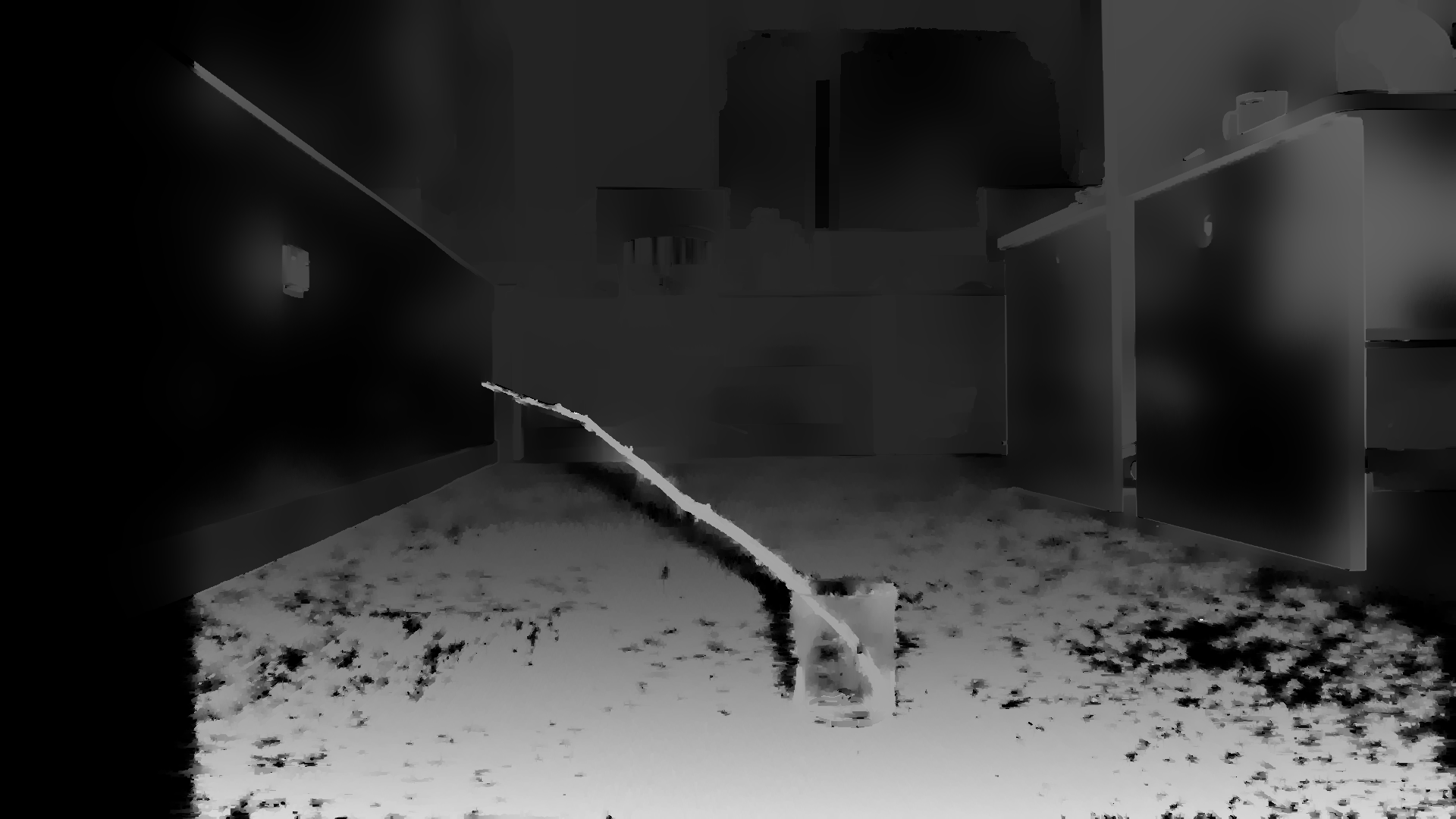}}
        \label{circle_image}
    \end{figure}

    \subsection{Performance Evaluation}

    System performance is comprehensively evaluated using established computer
    vision metrics:

    \textbf{Detection and Segmentation Metrics:}
    \begin{itemize}
        \item mAP$_{box50\text{--}95}$: Mean Average Precision for bounding box detection
            across IoU thresholds 0.5-0.95

        \item mAP$_{mask50\text{--}95}$: Mean Average Precision for pixel-level segmentation
            masks across IoU thresholds 0.5-0.95
    \end{itemize}

    \textbf{Depth Estimation Metrics:}
    \begin{itemize}
        \item Root Mean Square Error (RMSE) between estimated and ground truth depths
            for quantitative accuracy assessment
    \end{itemize}

    \begin{equation}
        \text{mAP}_{50\text{--}95}= \frac{1}{10}\sum_{t=0.5}^{0.95}\text{AP}(t) \label{eq:map}
    \end{equation}

    \begin{equation}
        \text{RMSE}= \sqrt{\frac{1}{n}\sum_{i=1}^{n}(y_{i}- \hat{y}_{i})^{2}}\label{eq:rmse}
    \end{equation}

    where $n$ is the number of samples, $y_{i}$ is the ground truth depth, and
    $\hat{y}_{i}$ is the estimated depth for sample $i$.

    \section{Experimental Results and Analysis}

    Comprehensive experimental evaluation demonstrates the effectiveness of our integrated
    approach and reveals significant architectural differences in branch detection
    capabilities. Our results establish clear performance hierarchies between
    different detection paradigms and conclusively validate the feasibility of
    stereo vision-based systems for precision forestry applications.

    \subsection{Branch Detection Performance}

    Table~\ref{Branches} presents comprehensive comparative performance results following
    standardized training protocols across all evaluated architectures. The
    results reveal substantial and consistent performance differences between
    detection paradigms.

    \begin{table}[h!]
        \centering
        \caption{Comparative Performance of Detection Architectures for Branch
        Segmentation. Results demonstrate YOLO's substantial superiority over Mask
        R-CNN variants across both detection and segmentation metrics, with
        YOLOv8s-seg achieving optimal balance of accuracy and efficiency.}
        \label{Branches}
        \begin{tabular}{lcc}
            \toprule \textbf{Model}    & \textbf{mAP$_{box}$} & \textbf{mAP$_{mask}$} \\
            \midrule Mask R-CNN R50-C4 & 76.86                & 0.06                  \\
            Mask R-CNN R50-DC5         & 77.54                & 9.16                  \\
            Mask R-CNN R50-FPN         & 79.19                & 6.75                  \\
            Mask R-CNN R101-C4         & 88.05                & 0.05                  \\
            Mask R-CNN R101-DC5        & 79.12                & 9.94                  \\
            Mask R-CNN R101-FPN        & 84.09                & 2.95                  \\
            Mask R-CNN X101-FPN        & 85.52                & 11.55                 \\
            \midrule YOLOv8n-seg       & 98.9                 & 77.4                  \\
            YOLOv8s-seg                & 99.5                 & 82.0                  \\
            YOLOv8m-seg                & 99.6                 & 81.6                  \\
            YOLOv8l-seg                & 99.2                 & 80.1                  \\
            YOLOv8x-seg                & 98.7                 & 77.1                  \\
            YOLOv9c-seg                & 98.9                 & 80.9                  \\
            YOLOv9e-seg                & 98.8                 & 80.0                  \\
            \bottomrule
        \end{tabular}
    \end{table}

    The experimental results reveal several critical findings with significant implications
    for autonomous forestry applications:

    \textbf{Detection Accuracy Analysis:} YOLO architectures consistently
    achieve exceptional detection performance with mAP$_{box}$ scores exceeding 98\%
    across all variants, compared to 76-88\% for Mask R-CNN implementations. This
    represents a substantial 10-22 percentage point improvement in detection accuracy.
    The superior performance stems from YOLO's unified architecture that
    processes the entire image simultaneously, enabling better contextual understanding
    of elongated branch structures within complex forest scenes.

    \textbf{Segmentation Performance Comparison:} The performance gap becomes
    even more pronounced in segmentation tasks, where precise pixel-level accuracy
    is essential for autonomous pruning applications. YOLO variants achieve mAP$_{mask}$
    scores of 77-82\%, while Mask R-CNN implementations remain below 12\% in
    most configurations. YOLOv8s-seg provides the optimal balance with 99.5\%
    box mAP and 82.0\% mask mAP, representing an 8-fold improvement over the
    best-performing Mask R-CNN variant.

    \textbf{Architectural Analysis:} The dramatic performance difference likely originates
    from fundamental architectural distinctions. Mask R-CNN's two-stage approach,
    originally designed for general object detection, appears poorly suited for
    elongated natural structures like tree branches. The region proposal stage may
    fragment long, thin branches into multiple disconnected segments, severely degrading
    segmentation quality. Conversely, YOLO's single-stage architecture with anchor-free
    detection naturally handles objects with extreme aspect ratios, making it
    inherently better suited for branch detection tasks.

    \subsection{Stereo Depth Estimation Results}

\begin{figure}[htbp]
        \centering
        \caption{Comprehensive depth estimation pipeline demonstrating
        systematic improvement through processing stages: (a,b) Original stereo pair
        with natural lighting variations, (c,d) Preprocessed images with
        enhanced contrast and noise reduction, (e) Raw SGBM disparity output showing
        initial depth estimates, (f) WLS-filtered disparity map with preserved
        edge structure and reduced noise artifacts.}
        \subfigure[Original Left]{\includegraphics[width=0.23\textwidth]{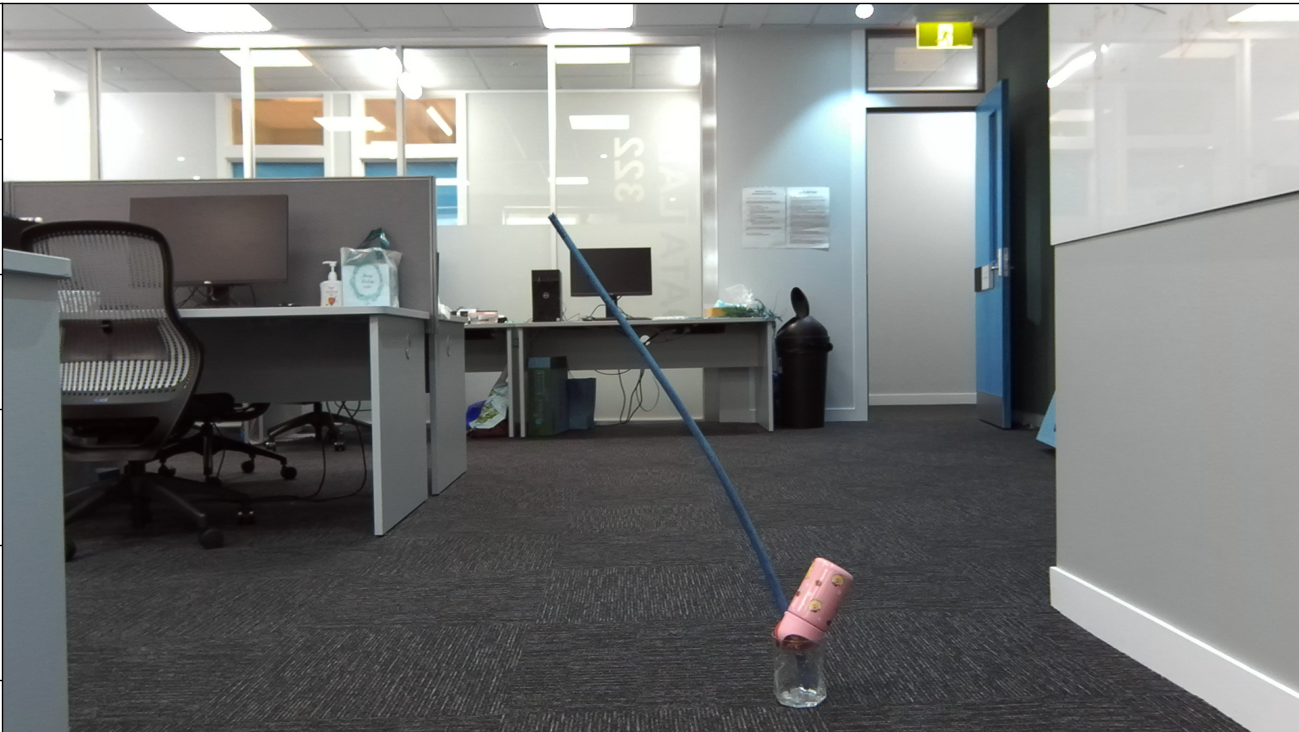}}
        \subfigure[Original Right]{\includegraphics[width=0.23\textwidth]{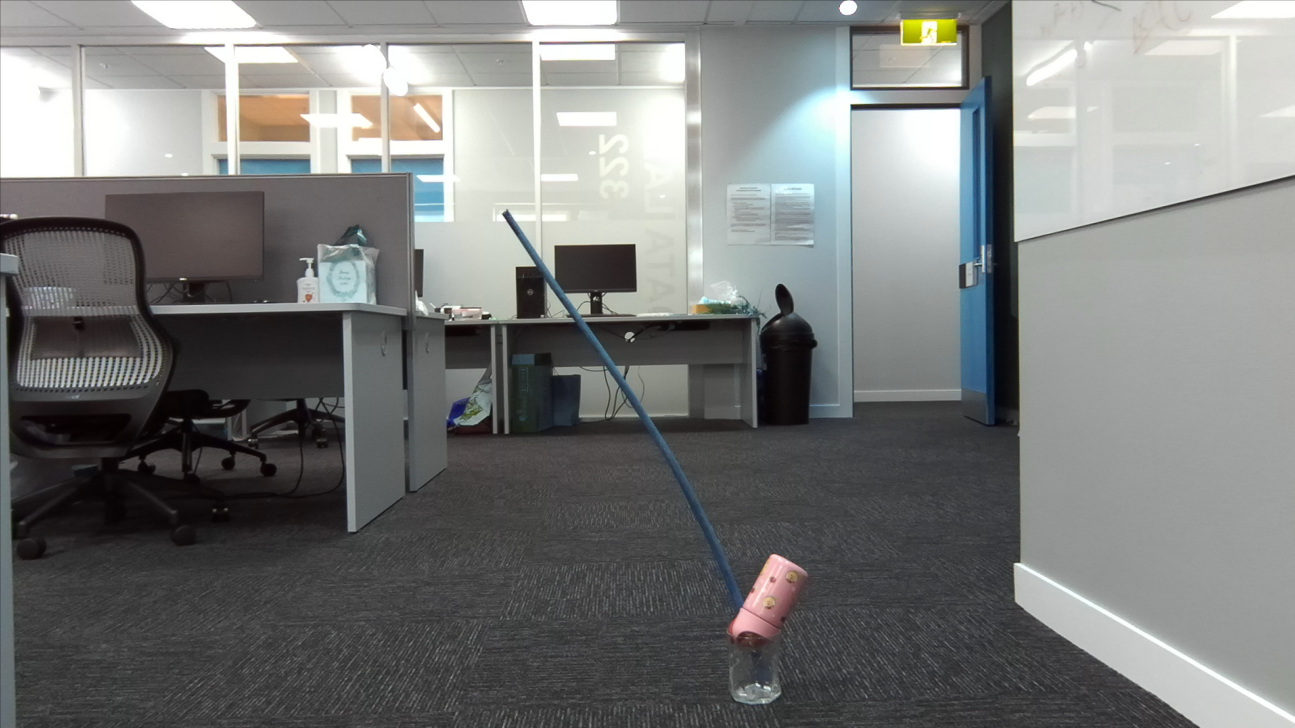}}
        \subfigure[Preprocessed Left]{\includegraphics[width=0.23\textwidth]{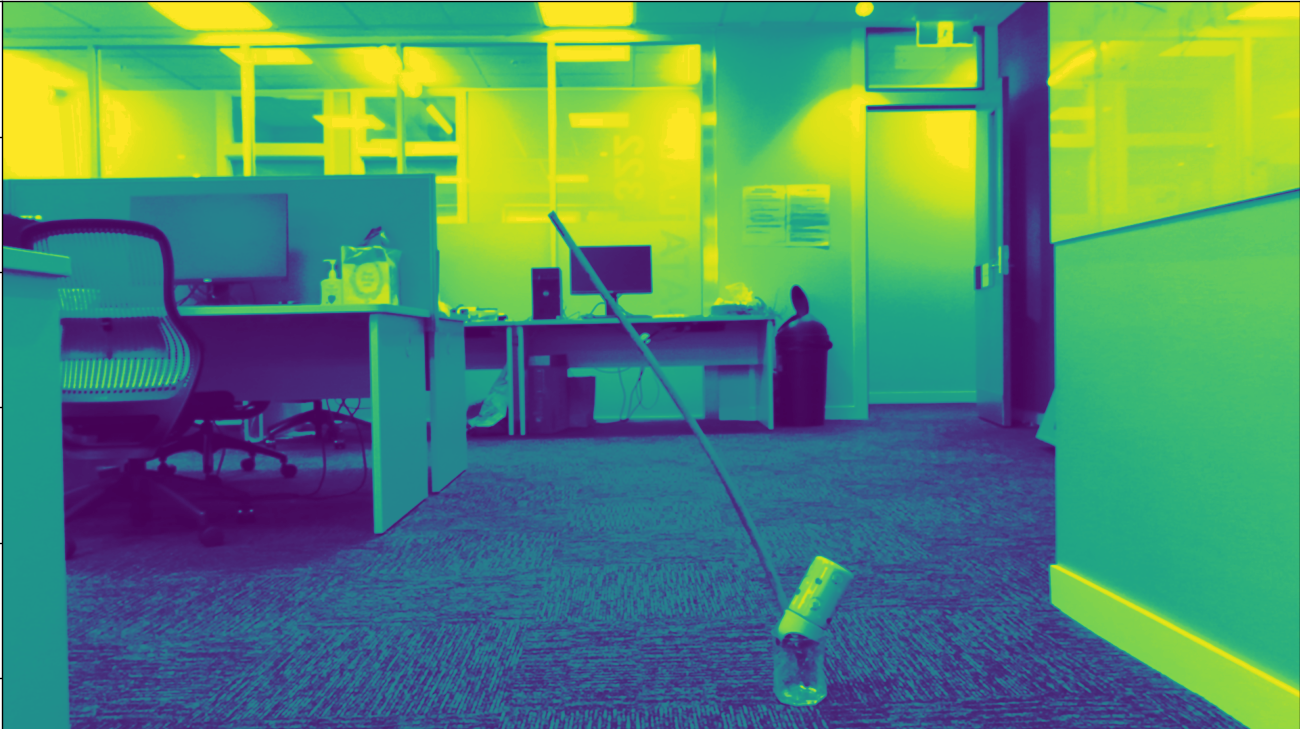}}
        \subfigure[Preprocessed Right]{\includegraphics[width=0.23\textwidth]{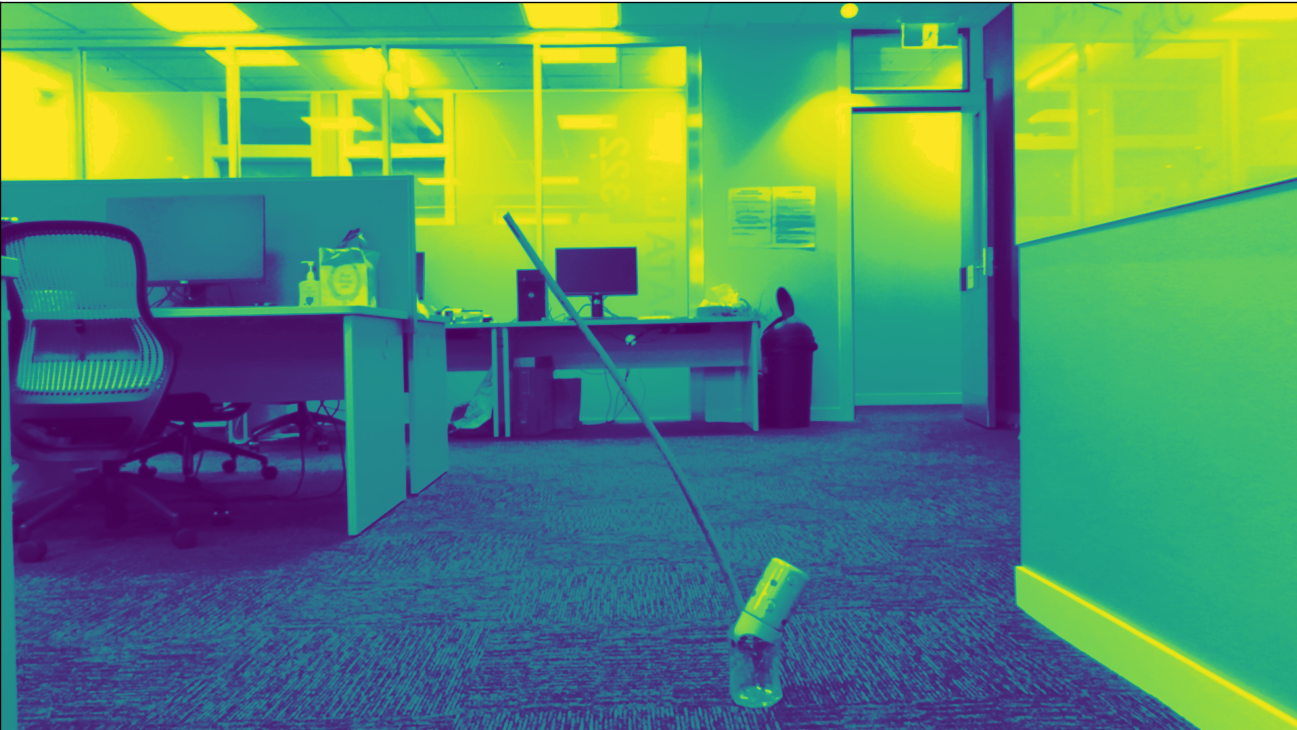}}
        \subfigure[Raw SGBM]{\includegraphics[width=0.23\textwidth]{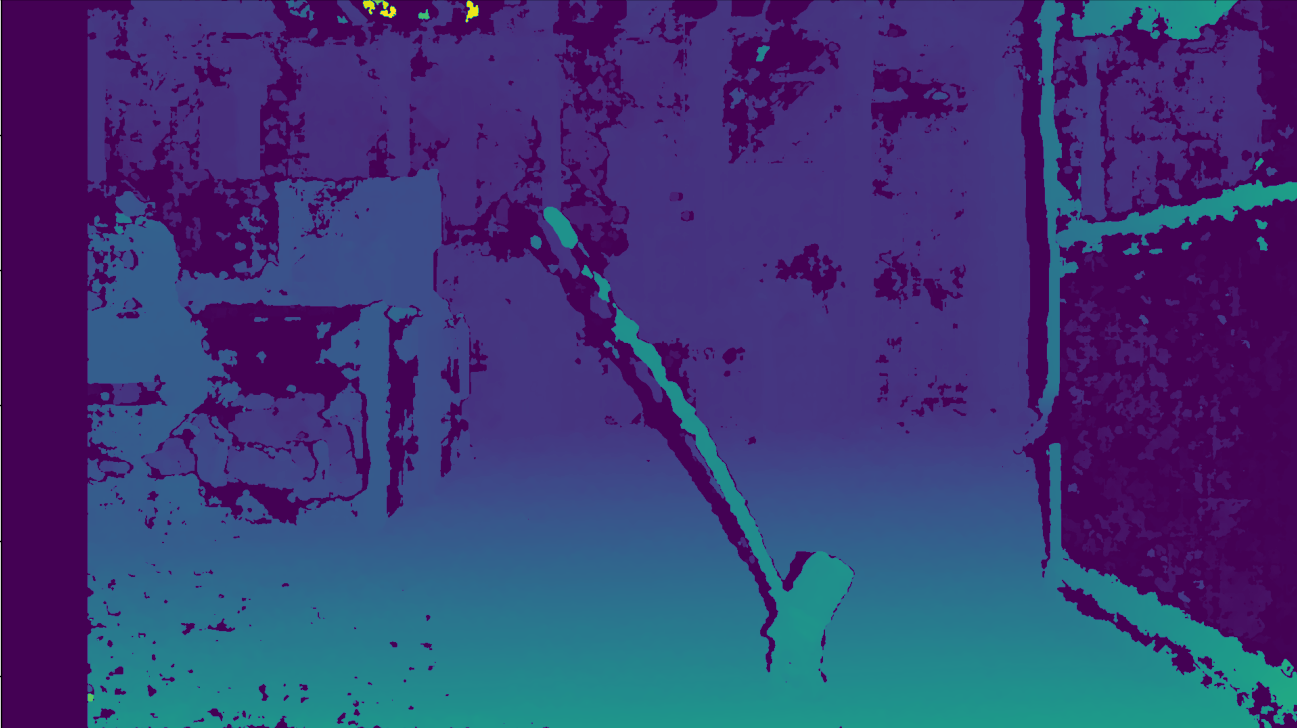}} \subfigure[WLS
        Filtered]{\includegraphics[width=0.23\textwidth]{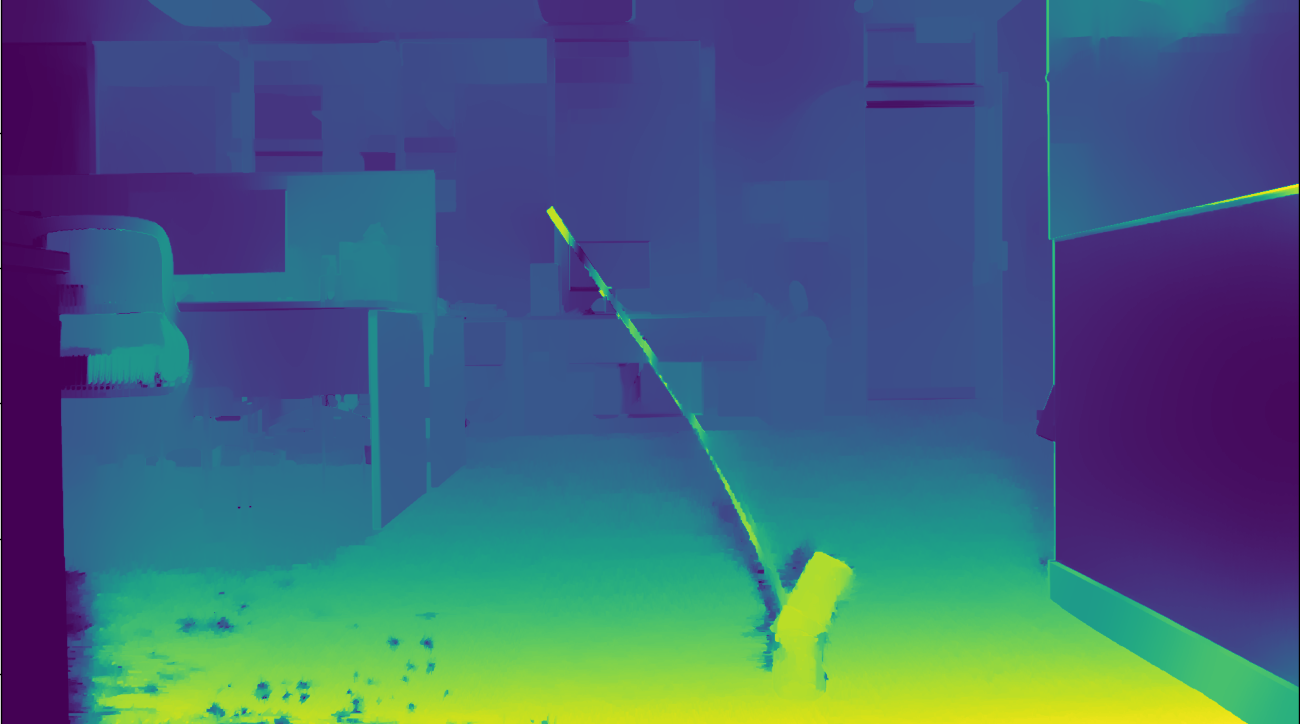}} \label{Traditional}
    \end{figure}

    Figure~\ref{Traditional} demonstrates the comprehensive depth estimation pipeline
    progression from raw stereo inputs through preprocessing stages to refined disparity
    maps suitable for precise 3D localization.

    The depth estimation pipeline demonstrates systematic quality improvement
    through each processing stage:

    \textbf{Preprocessing Enhancement:} Image preprocessing significantly improves
    stereo matching robustness through noise reduction, contrast optimization, and
    illumination normalization. This preprocessing stage is particularly
    beneficial in forest environments with challenging and variable lighting
    conditions that can compromise raw stereo matching performance.

    \textbf{SGBM Performance:} Raw SGBM output provides reliable depth estimates
    across most image regions, with occasional noise artifacts limited to
    texture-poor homogeneous areas where correspondence matching faces inherent
    challenges.

    \textbf{WLS Filtering:} Weighted Least Squares filtering dramatically improves
    disparity quality by effectively eliminating noise artifacts while precisely
    preserving edge information at critical branch boundaries. This ensures
    sharp depth discontinuities that are crucial for accurate branch
    segmentation and subsequent 3D localization.

    \textbf{Quantitative Assessment:} Evaluation across the operational range yields
    predictable accuracy characteristics that closely follow theoretical stereo geometry
    expectations. The progressive error increase at greater ranges follows the
    quadratic relationship defined in Equation~\ref{eq:stereo_depth}, yet achieved
    accuracy remains consistently within acceptable tolerances for autonomous pruning
    operations.

    \textbf{Environmental Robustness:} The system demonstrates consistent
    performance across diverse environmental conditions including direct
    sunlight, overcast skies, and artificial illumination. Accuracy degradation remains
    minimal even under challenging lighting conditions, indicating robust
    operational suitability for real-world forestry applications.

    \subsection{Integrated System Performance}

\begin{figure}[htbp]
        \centering
        \caption{Integrated system performance analysis across operational
        ranges: (a--c) Depth maps at 1 m, 1.5 m, and 2 m distances showing
        consistent branch detection and depth estimation, (d--f) Corresponding
        depth measurement distributions demonstrating precision characteristics
        and statistical reliability for autonomous pruning applications.}
        \subfigure[Depth Map 1 m]{\includegraphics[width=0.23\textwidth]{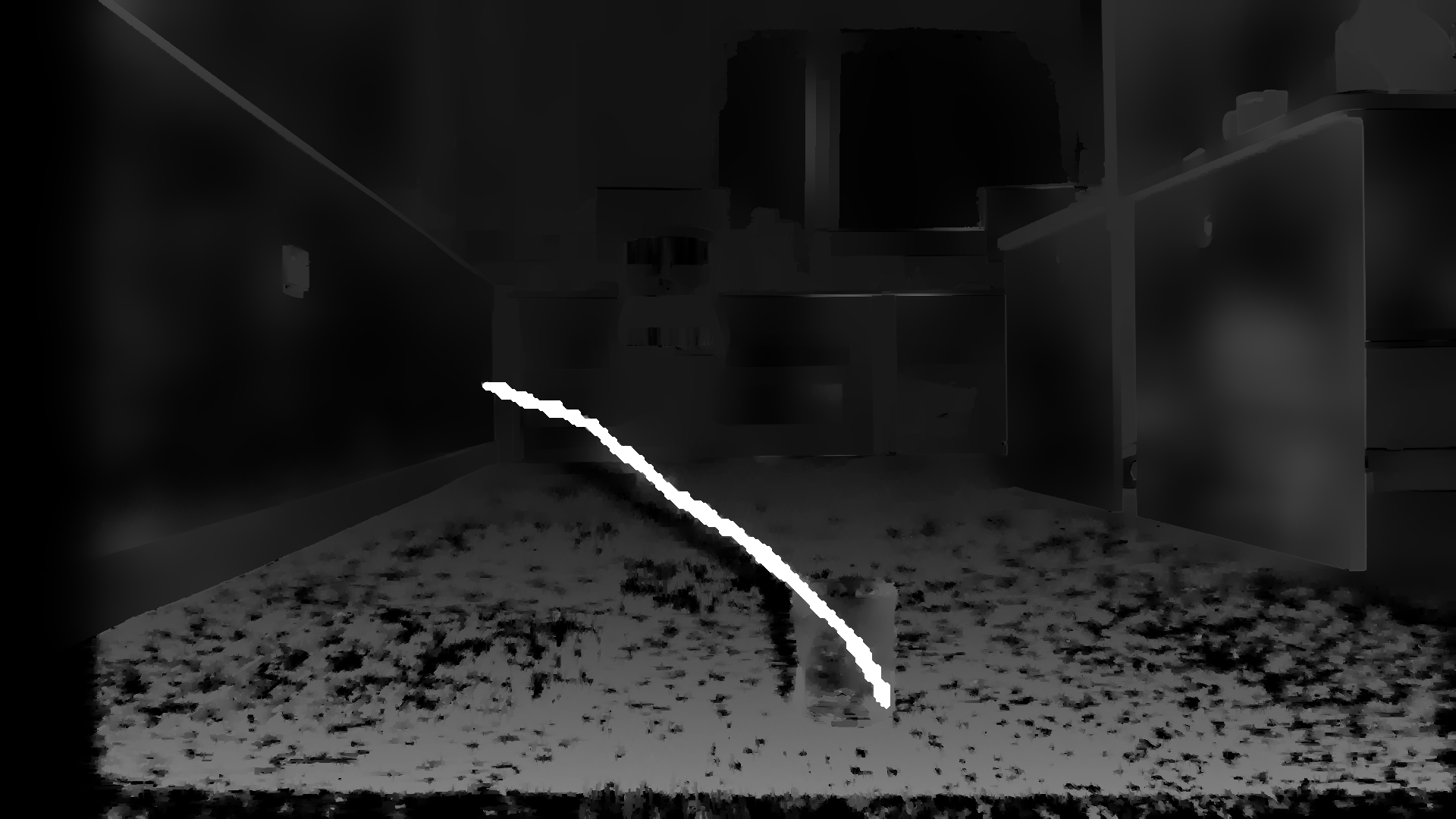}}
        \subfigure[Depth Map 1.5 m]{\includegraphics[width=0.23\textwidth]{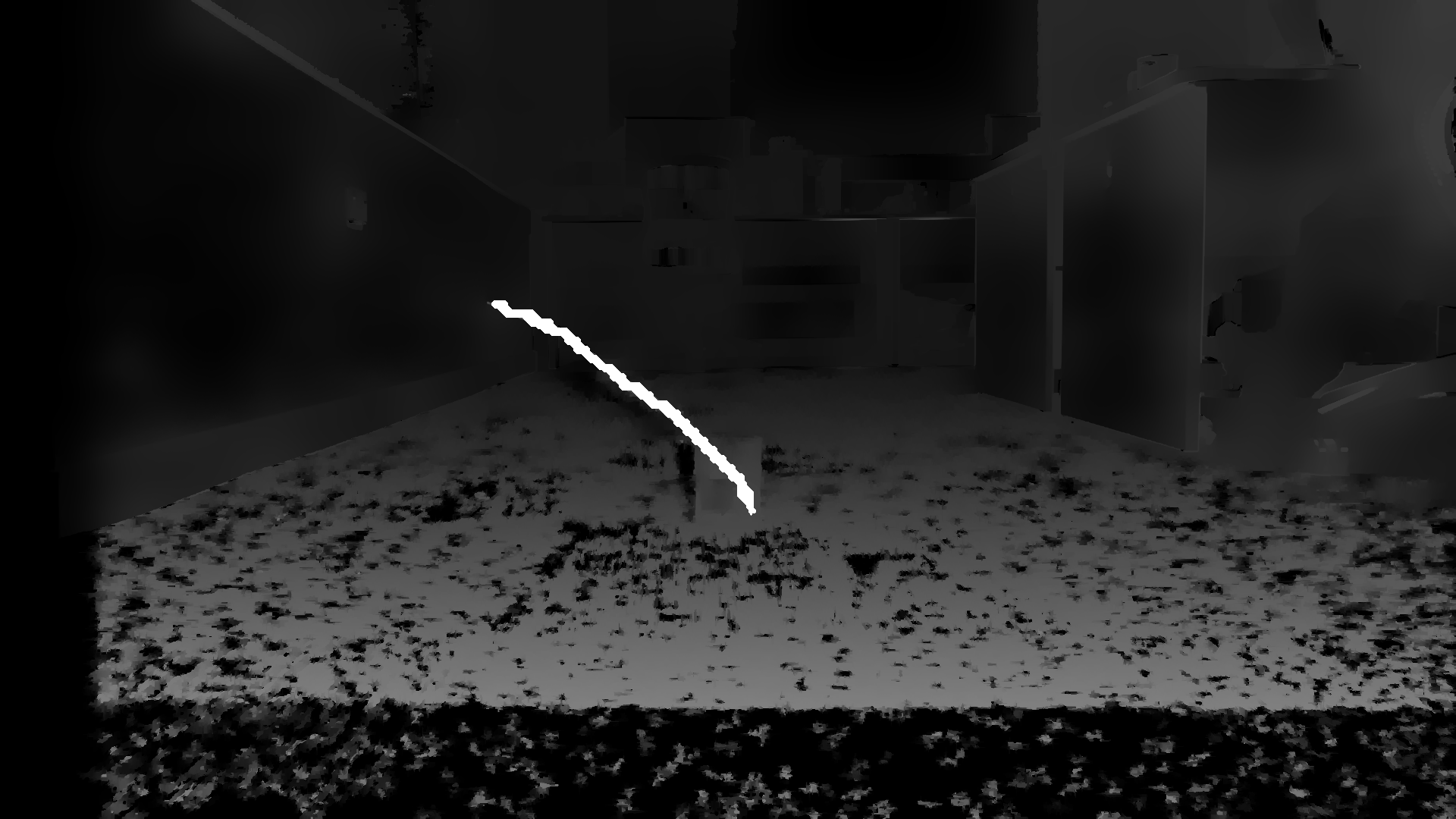}}
        \subfigure[Depth Map 2 m]{\includegraphics[width=0.23\textwidth]{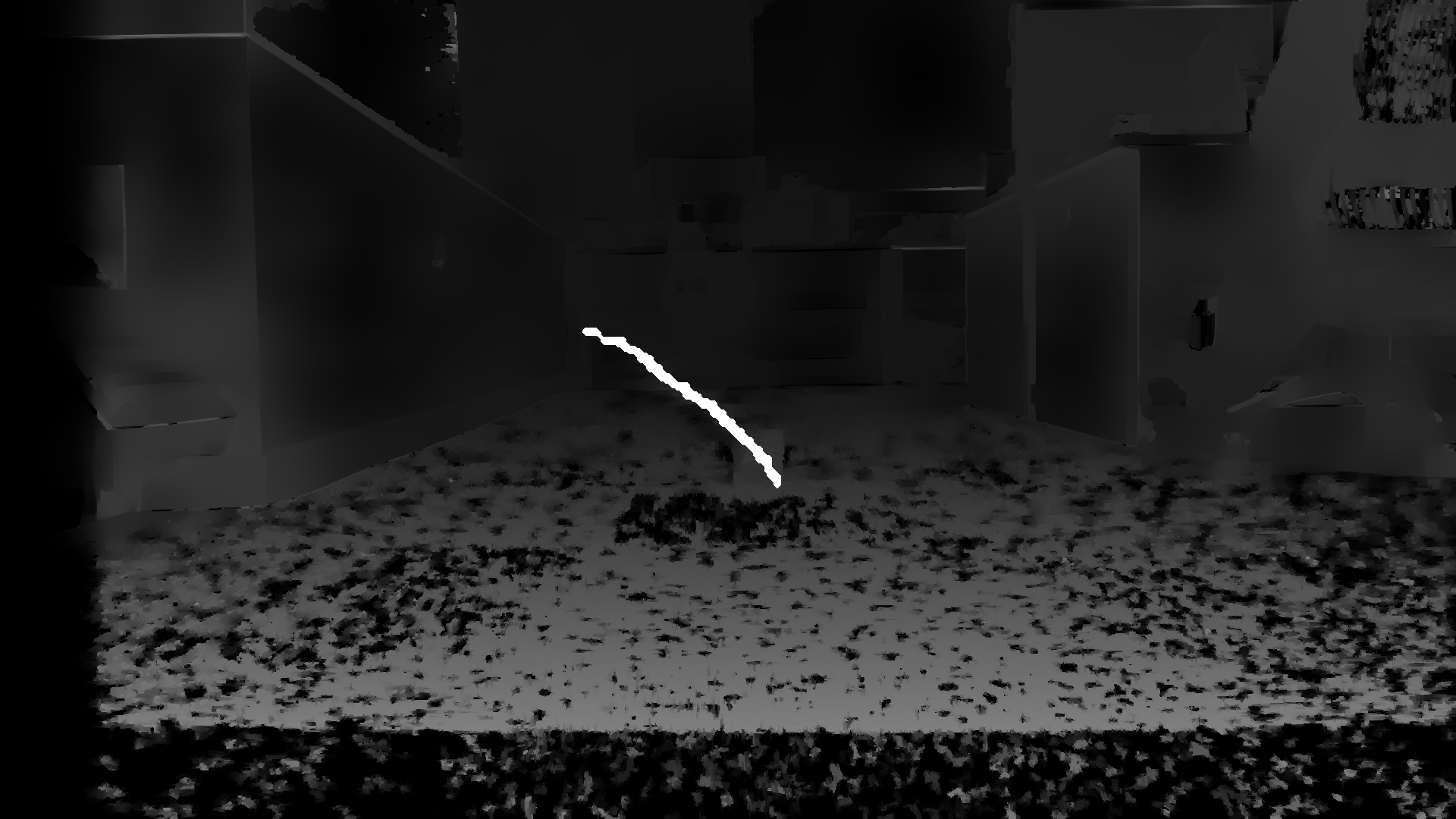}}
        \subfigure[Distribution 1 m]{\includegraphics[width=0.23\textwidth]{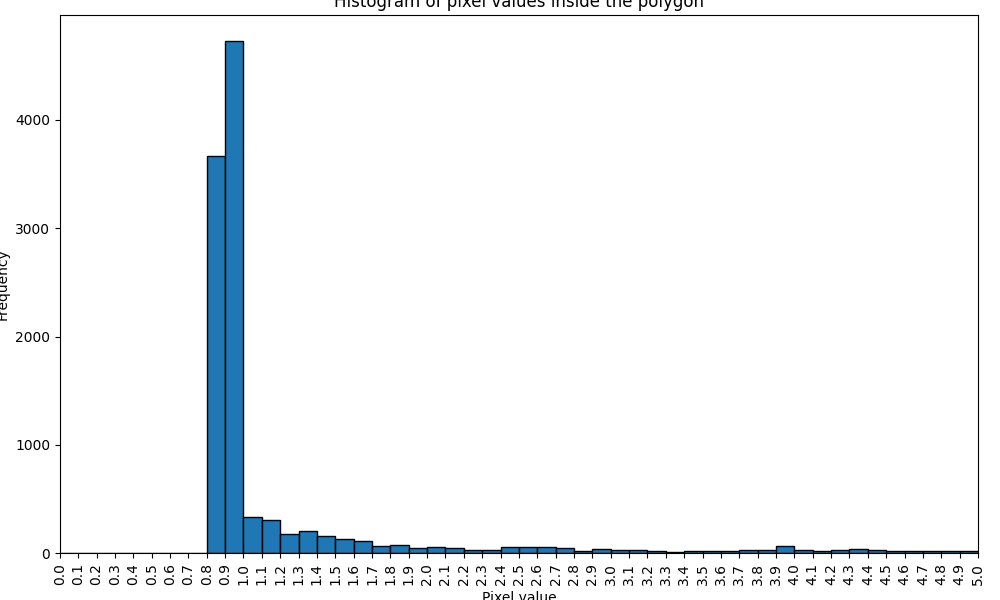}}
        \subfigure[Distribution 1.5 m]{\includegraphics[width=0.23\textwidth]{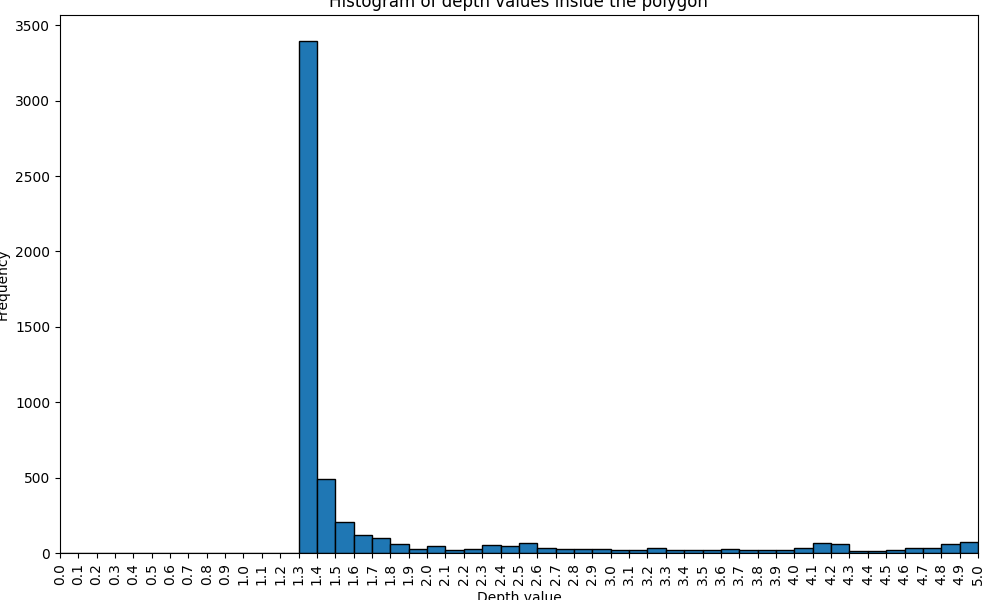}}
        \subfigure[Distribution 2 m]{\includegraphics[width=0.23\textwidth]{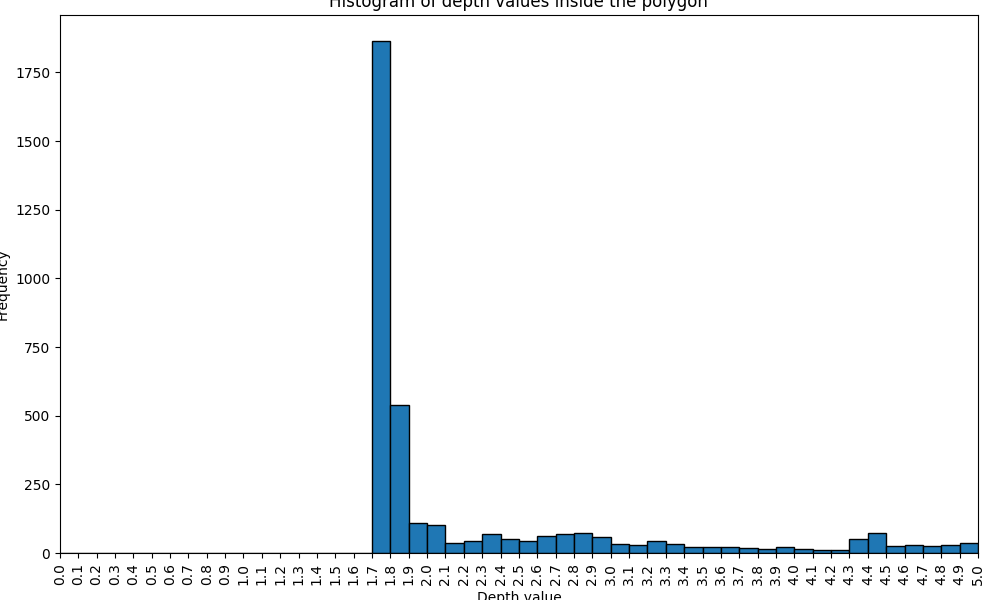}}
        \label{SGBM_and_distribution}
    \end{figure}

    Figure~\ref{SGBM_and_distribution} illustrates comprehensive system performance
    combining branch detection with depth estimation across the operational range,
    demonstrating both accuracy and consistency requirements for autonomous
    applications.

    Integrated system evaluation reveals several key performance characteristics
    essential for autonomous forestry applications:

    \textbf{Operational Range Capability:} The system successfully detects and
    localizes branches across the complete operational range required for
    autonomous pruning systems within a 2-meter working distance. Detection accuracy
    remains consistently high at all tested distances, while 3D localization
    errors stay within acceptable tolerances even at maximum operational range.

    \textbf{Measurement Precision and Consistency:} Depth distributions exhibit clear,
    well-defined peaks at true branch distances with standard deviations
    increasing progressively at greater ranges. The distributions become wider at
    extended distances, consistent with fundamental stereo vision theoretical
    limitations, while maintaining practical accuracy sufficient for autonomous navigation
    and manipulation tasks.

    \textbf{Processing Efficiency and Real-Time Performance:} Complete pipeline
    execution approaches real-time performance requirements on modern GPU
    hardware, suitable for autonomous navigation systems. Memory consumption
    remains within reasonable bounds for embedded deployment on high-performance
    drone platforms equipped with advanced computing units.

    \textbf{Robustness Analysis:} System performance remains stable across environmental
    variations including lighting changes, viewing angles, and branch orientations.
    Detection and depth estimation accuracy show minimal degradation under
    challenging conditions, indicating robust performance characteristics suitable
    for field deployment.

    \textbf{Scalability Assessment:} The system successfully handles complex
    scenes containing multiple branches without significant performance
    degradation. Processing time scales sub-linearly with branch count, showing
    minimal increase for complex multi-branch scenes compared to single-branch
    scenarios.

    \section{Discussion}

    This research demonstrates the viability of cost-effective computer vision solutions
    for autonomous forestry applications through strategic integration of modern
    deep learning detection techniques with classical stereo vision methods. Our
    findings have significant implications for both the technical development of
    autonomous systems and the economic transformation of commercial forestry
    operations.

    \subsection{Technical Contributions and Insights}

    \textbf{YOLO Superiority for Branch Detection:} Our evaluation demonstrates
    YOLO's significant advantages over Mask R-CNN for branch segmentation. YOLO's
    single-stage architecture maintains spatial continuity in elongated objects,
    unlike two-stage detectors that may fragment thin branches into disconnected
    segments.

    \textbf{Stereo Vision Effectiveness:} The integration of SGBM with WLS filtering
    proves effective for forest environments, achieving sufficient accuracy for
    precision forestry applications while providing a cost-effective alternative
    to LiDAR systems.

    \textbf{System Integration:} The successful fusion of detection and depth estimation
    demonstrates effective spatial registration through statistical depth measures
    extracted from segmented regions.

    \subsection{Economic and Practical Implications}

    \textbf{Cost Effectiveness:} Stereo vision provides dense depth mapping at significantly
    lower cost than multi-beam LiDAR systems, making autonomous pruning
    economically viable for broader forestry operations.

    \textbf{Safety Benefits:} The system reduces manual exposure to height-related
    hazards by enabling remote branch assessment and localization.

    \textbf{Deployment Feasibility:} Processing performance approaches real-time
    requirements suitable for semi-autonomous operations, with memory requirements
    compatible with modern embedded platforms.
    \subsection{Current Limitations}

    \textbf{Processing Speed:} Current processing latency may require
    optimization for fully autonomous operation in dynamic environments.

    \textbf{Environmental Conditions:} Validation under extreme weather conditions
    and varying environmental factors requires further investigation.

    \textbf{Operational Range:} The 2-meter working distance may limit
    applicability to operations requiring greater ranges.

    \textbf{Dataset Scope:} Current focus on radiata pine under controlled conditions
    limits generalization to diverse species and natural environments.

    \subsection{Future Research Directions}

    Key areas for future development include:

    \begin{enumerate}
        \item \textbf{Performance Optimization:} Model compression and hardware
            acceleration to achieve real-time operation

        \item \textbf{Range Extension:} Hybrid sensing approaches for extended
            operational ranges

        \item \textbf{Environmental Validation:} Field testing under diverse
            weather and forest conditions

        \item \textbf{Multi-Species Adaptation:} Extension to different tree
            species and forest types

        \item \textbf{System Integration:} Development of complete autonomous
            pruning systems
    \end{enumerate}

    \section{Conclusions}

    This research demonstrates the feasibility of integrating advanced object
    detection with stereo vision for autonomous tree branch detection and
    localization in forestry applications. The proposed system addresses critical
    safety and economic challenges while providing a cost-effective alternative to
    traditional methods.

    \subsection{Key Contributions}

    \begin{itemize}
        \item \textbf{Branch Detection:} YOLO architectures demonstrate superior
            performance over Mask R-CNN for elongated natural structures like
            tree branches.

        \item \textbf{Depth Estimation:} SGBM with WLS filtering provides
            reliable depth estimation using affordable stereo cameras as an
            alternative to expensive LiDAR sensors.

        \item \textbf{System Integration:} The integrated system achieves near
            real-time performance suitable for autonomous forestry applications.
    \end{itemize}

    \subsection{Practical Implications}

    The system enhances worker safety by reducing exposure to hazardous
    operations, improves efficiency through automation, and provides cost-effective
    solutions using consumer-grade hardware.

    \subsection{Future Directions}

    Future work will focus on processing optimization, range extension, and comprehensive
    validation under diverse environmental conditions to enable production-ready
    autonomous pruning systems.
    \section{Acknowledgement}
    This work was supported in part by the MBIE Endeavor Research Programme UOCX2104, the MBIE Data Science SSIF Fund under contract RTVU1914, and Marsden Fund of New Zealand Government under Contracts VUW2115.

    \bibliographystyle{IEEEtran}
    \bibliography{references}
\end{document}